\documentclass[letterpaper]{article} 
\usepackage{aaai24}  
\usepackage{times}  
\usepackage{helvet}  
\usepackage{courier}  
\usepackage[hyphens]{url}  
\usepackage{graphicx} 
\usepackage{natbib}  
\usepackage{caption} 
\usepackage{algorithm}
\usepackage{amsmath}
\usepackage{amssymb}
\usepackage{booktabs}
\usepackage{colortbl}
\usepackage{bbding}

\definecolor{LightCyan}{rgb}{0.88,1,1}
\graphicspath{{figs/}}
\newcommand{\etal}{\textit{et al}.}
\newcommand{\ie}{\textit{i}.\textit{e}.~}
\newcommand{\eg}{\textit{e}.\textit{g}.~}

\usepackage{newfloat}
\usepackage{listings}
\DeclareCaptionStyle{ruled}{labelfont=normalfont,labelsep=colon,strut=off} 
\floatstyle{ruled}
\newfloat{listing}{tb}{lst}{}
\floatname{listing}{Listing}
\nocopyright

\title{Category Feature Transformer for Semantic Segmentation}
\author{
    Quan Tang\textsuperscript{\rm 1,2}\thanks{Work done during an internship at Huawei Noah’s Ark Lab.},
    Chuanjian Liu\textsuperscript{\rm 2},
    Fagui Liu\textsuperscript{\rm 1,3\textdagger},
    Yifan Liu\textsuperscript{\rm 4},\\
    Jun Jiang\textsuperscript{\rm 1},
	Bowen Zhang\textsuperscript{\rm 4},
	Kai Han\textsuperscript{\rm 2},
	Yunhe Wang\textsuperscript{\rm 2}\thanks{Corresponding author.}
}
\affiliations{
    \textsuperscript{\rm 1}South China University of Technology~
    \textsuperscript{\rm 2}Huawei Noah’s Ark Lab~
    \textsuperscript{\rm 3}Peng Cheng Lab~
    \textsuperscript{\rm 4}The University of Adelaide\\
    \{csquantang, csjun.jiang\}@mail.scut.edu.cn,
    \{yifan.liu04, b.zhang\}@adelaide.edu.au\\
    fgliu@scut.edu.cn,
    \{liuchuanjian, kai.han, yunhe.wang\}@huawei.com
}

\begin{document}

\maketitle

\begin{abstract}
Aggregation of multi-stage features has been revealed to play a significant role in semantic segmentation. Unlike previous methods employing point-wise summation or concatenation for feature aggregation, this study proposes the Category Feature Transformer (CFT) that explores the flow of category embedding and transformation among multi-stage features through the prevalent multi-head attention mechanism. CFT learns unified feature embeddings for individual semantic categories from high-level features during each aggregation process and dynamically broadcasts them to high-resolution features. Integrating the proposed CFT into a typical feature pyramid structure exhibits superior performance over a broad range of backbone networks. We conduct extensive experiments on popular semantic segmentation benchmarks. Specifically, the proposed CFT obtains a compelling 55.1\% mIoU with greatly reduced model parameters and computations on the challenging ADE20K dataset.
\end{abstract}

\section{Introduction}
\label{sec:1}
Semantic segmentation, a primary yet challenging perception task in computer vision, associates every pixel with a pre-defined semantic category. It produces dense predictions that match the resolution of the input image, where meaningful regions naturally emerge at the scene level. Deep learning methods have significantly aided the advancement of this field, with the aggregation of multi-stage features being critical in enhancing recognition accuracy.

\begin{figure}[t]
    \centering
    \includegraphics[width=.98\linewidth]{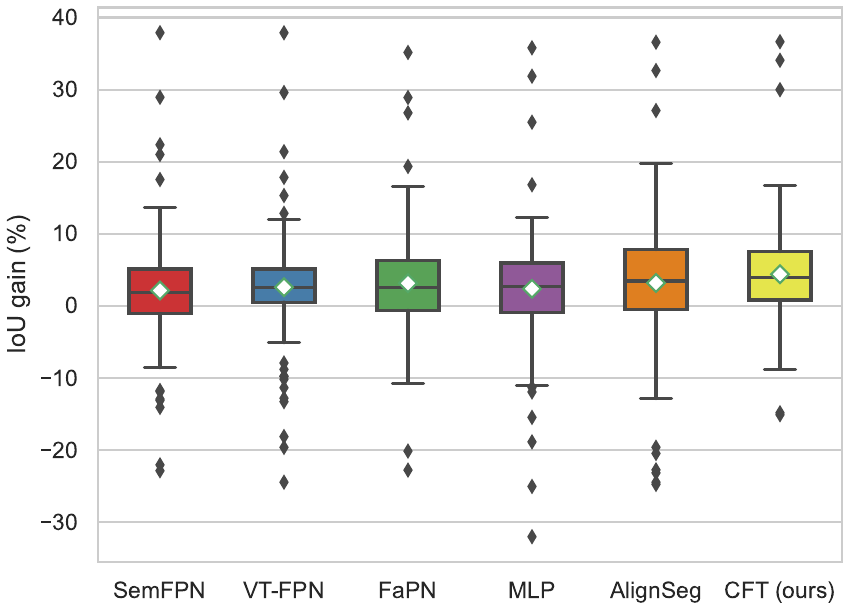}
    \caption{Numerical distribution in boxplot of per-category accuracy gain on the ADE20K dataset with different multi-stage feature aggregation methods. Results suggest that our CFT achieves the highest mIoU gain (see white rhombuses) while notably alleviates the accuracy drop on minority categories (the below black dots), denoting a more consistent accuracy improvement over all categories.}
    \label{fig:1}
\end{figure}

Typically, a backbone network functioning as an encoder yields multi-stage features with varying scales. A common practice starts from features at the last stage and then performs progressive feature upsampling and aggregates from the previous stage until all features in all stages are considerably utilized, \eg the feature pyramid network (FPN)~\cite{lin2017feature}. Going one step further, some studies~\cite{huang2021fapn,huang2022alignseg} focus on the context misalignment problem between features of adjacent stages and try to align these features before element-wise aggregation. In this study, however, we observe that while these paradigms improve the overall scene parsing accuracy, they lead to significant differences in the performance gain for each semantic category. As shown in Fig.~\ref{fig:1}, the recognition accuracy for some categories incurs a sharp drop after aggregating multi-stage features. Thus exploring a robust multi-stage feature aggregation method is still imperative.

Recently, the Transformer~\cite{vaswani2017attention}, originating from natural language processing, has been successfully adapted to the field of computer vision, known as Vision Transformer (ViT)~\cite{dosovitskiy2020image}. The multi-head attention mechanism is a principal component that generates content-dependent attention weights for aggregating any-distance features in multiple subspaces. Currently, vision transformers have two main applications in semantic segmentation. They can either work as a general backbone to generate enhanced feature representation~\cite{zheng2021rethinking} or learn dynamic filters in the decoding process to produce segmentation masks by properly operating the sequence-to-sequence characteristic~\cite{lin2023structtoken}. Differently, this study takes a fresh angle and explores the capacity of multi-head attention in aggregating multi-stage features due to its dynamic and unique global-modeling ability.

An intuition to implement this idea is to aggregate features with the previous stage as query and the current stage as key/value. In this way, two-stage features are dynamically aggregated without explicit bilinear upsampling operation, thus avoiding injecting semantic noise due to the limited feature reconstruction capability~\cite{tian2019decoders}. Because of the vast resolutions of early-stage feature maps, this naive implementation suffers from unendurable computational costs. Based on the assumption that pixels within a neighbor region are likely of the same semantic category and have comparable features, prior methods~\cite{xie2021segformer} downsample the key/value to gather information in a rectangular pooling window (\ie regular grid), which lightens the computational burden. We argue that by this operation, the resulting key/value features carry jumbly semantics that hinders feature aggregation because regular grids just approximate (not follow) the semantic structure in the real world.

To this end, this study proposes Category Feature Transformer (CFT) for an effective top-down feature aggregation in semantic segmentation. Based on the current high-level feature, CFT explicitly learns category-specific masks to define multiple \emph{irregular} regions where each corresponds to an individual semantic category. Then unified category feature embeddings that carry highly consistent semantics are obtained via averaging feature values in the corresponding irregular region. These category-specific features are then adaptively broadcast to the corresponding high-resolution (\ie low-level) feature map via a weighted-summation manner. With the help of this novel design, CFT achieves consistent accuracy improvement over three challenging benchmarks with notably reduced parameters and computations compared to other contemporary works. We conclude the main contributions as follows:
\begin{itemize}
	\item We revisit the multi-stage feature aggregation regime and observe that minority categories suffer from a severe accuracy drop after aggregating multi-stage features.
    \item We propose a novel scheme that learns category-specific feature embeddings in the context of multi-head cross-attention for the top-down feature aggregation paradigm in semantic segmentation. It guarantees highly consistent semantics and significantly reduces model complexity.
	\item We conduct extensive experimental studies on challenging benchmarks. Quantitative results suggest the effectiveness and ablation experiments provide insights into the properties. Specifically, CFT obtains a compelling 55.1\% mIoU on the ADE20K dataset.
\end{itemize}

\section{Related Work}
\label{sec:2}
\subsection{Encoder-Decoder}
In the last decade, deep learning techniques have greatly facilitated the development of semantic image segmentation. The fully convolutional network (FCN)~\cite{long2015fully} is a seminal work that formulates the task as a dense pixel classification problem. Since then, semantic segmentation networks have generally followed a pervasive encoder-decoder structure. The encoder produces multi-stage image features, and the decoder enriches the representation and recovers the feature resolution. As image classification networks are commonly utilized as the encoder, more emphasis has been placed on creating a unique decoder design. A popular paradigm employs context modules to learn multi-scale features with the help of the atrous convolution to maintain feature resolution, \eg the PSPNet~\cite{zhao2017pyramid} and DeepLab series~\cite{chen2017rethinking,chen2018encoder}. Another line aims to fully utilize these multi-stage features conveyed by the encoder. Both top-down aggregation~\cite{kirillov2019panoptic,huang2021fapn} and bottom-up fusion~\cite{pang2019towards,huang2022alignseg} with a parallel branch are developed, showing that aggregation of multi-stage features continuously improves the segmentation accuracy. We suggest that existing methods lead to severe performance degradation for some categories after aggregating multi-stage features and present a robust strategy based on category attention and cross-attention.

\subsection{Vision Transformer}
Vision Transformer (ViT)~\cite{dosovitskiy2020image} splits an image into multiple non-overlapping patches that are linearly transformed into image tokens. An extra classification token is then appended for final image classification. ViT sheds light on a different path that relying on conventional convolutional neural networks (CNN) seems unnecessary. In semantic segmentation, ViT-based methods have achieved advances in popular benchmarks. SETR~\cite{zheng2021rethinking} uses a standard ViT as the encoder to enhance feature representation. Segmenter~\cite{strudel2021segmenter} and StructToken~\cite{lin2023structtoken} present learnable tokens or structure prior to generate dynamic classification filters. Segformer~\cite{xie2021segformer} incorporates multi-head attention and convolution to obtain a position-encoding-free network. This study explores the capacity of multi-head attention in dynamically aggregating multi-stage features. We present a module to obtain a unified feature embedding for a given semantic category, which helps learn consistent semantics and reduce computational costs.

\subsection{Relations to Prior Works}
It's worth noting that both category embedding and cross-attention are extensive areas of research. Here we explicitly present conceptual differences between CFT and prior works. ACFNet~\cite{zhang2019acfnet} and OCRNet~\cite{yuan2020object} use category embedding as attention weights to reweigh pixel features to obtain pixel-region relations by building context modules based on a single encoder stage. CondNet~\cite{yu2021condnet} adopts it to learn a dynamic convolutional kernel to address intra-class distinction. In contrast, CFT focuses on multi-stage feature fusion and adapts category embedding directly as the key/value in cross-attention, transmitting category information accordingly to a high-resolution low-level feature map (\ie query) without an explicit noisy upsampling. Most recently, SegViT~\cite{zhang2022segvit} and MaskFormer series~\cite{cheng2021per,cheng2022masked} explore the capacity of transformer decoder in semantic segmentation. The conceptual difference lies in that CFT uses a low-level feature map as the query and directly aggregates category prior represented by key/value that is learned from high-level features, while the other two assume category prior in each learnable query and decode it successively by attending to pixel features. The assumption in query/key/value, as well as the feature interaction mode, is fundamentally different. Therefore, developing a potent FPN variant based on cross-attention and category attention is a \emph{non-trivial} work.

\section{Method}
\label{sec:3}
This section describes in detail the proposed CFT framework dedicated to providing a robust solution for aggregating multi-stage features in semantic segmentation. We begin with a glance at the overall architecture and move on to detailed designs of the method.

\subsection{Overview}
\label{sec:3-1}
\begin{figure}[t]
    \centering
    \includegraphics[width=0.96\linewidth]{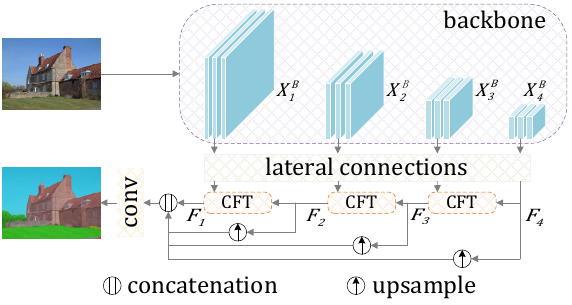}
    \caption{Overview of the framework, consisting of a backbone network to obtain multi-stage features and CFT to effectively aggregate these features. The aggregated features are then concatenated together after scale alignment for final pixel classification.}
    \label{fig:2}
\end{figure}

The overall architecture of the proposed framework is illustrated in Fig.~\ref{fig:2}. A backbone network is adopted to obtain multi-stage image features that are then aggregated by CFT in a top-down manner. Those multi-stage features from the encoder are first mapped to the same feature space by separate lateral/skip connections as they are of distinct channel dimensions. Finally, we upsample these aggregated features and concatenate them at the channel dimension for dense prediction. Note that the backbone is not limited to CNN or vision transformers that produce different-scale features. Without generality, this study employs ResNet~\cite{he2016deep}, MiT~\cite{xie2021segformer} and SwinT~\cite{liu2021swin} as the backbone network.

\subsection{Category Feature Transformer}
\label{sec:3-2}
\begin{figure*}[t]
    \centering
    \includegraphics[width=0.98\linewidth]{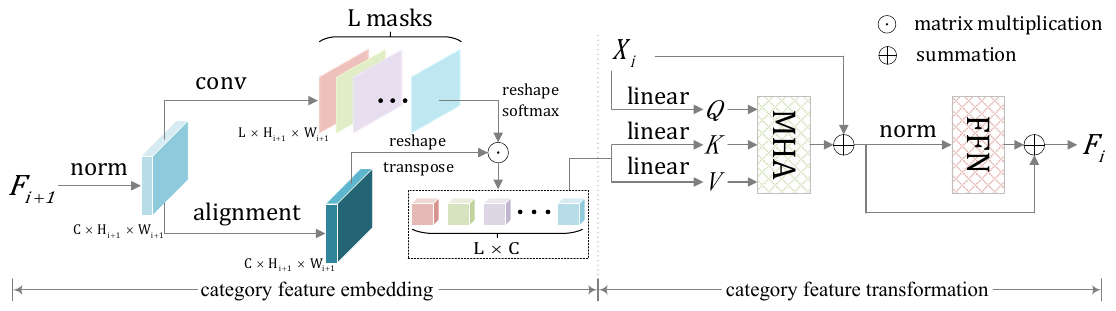}
    \caption{A CFT block, comprising category feature embedding and transformation. MHA is short for the multi-head attention and FFN feed-forward network. $L$ is the number of semantic categories. We omit some reshaping operations for simplicity.}
    \label{fig:3}
\end{figure*}

CFT works as the critical component for an efficacious fusion of adjacent-stage features, comprising category feature embedding and transformation, as shown in Fig.~\ref{fig:3}. Category feature embedding explicitly learns a unified feature embedding for each semantic category based on the high-level feature, which on the one hand, obtains consistent semantics and, on the other hand, dramatically reduces computational complexity. Category feature transformation computes the affinity scores between each low-level feature dot and the learned category embeddings. These affinity scores serve as attention weights that dynamically broadcast high-level semantics to high-resolution features, where no upsampling operator is used, \eg bilinear upsampling. By avoiding the use of \emph{explicit} bilinear upsampling and point-wise aggregation operators such as summation or concatenation, we can reduce the introduction of semantic noise.

Let $x \in \mathbb{R}^{3 \times H \times W}$ be the input image, where $H$ and $W$ represent the height and width, respectively. Then we obtain multi-stage image features $\{X_i^B\}_{i=1}^4$ using the hierarchical backbone network. $X_i^B \in \mathbb{R}^{C_i \times H_i \times W_i}$ is the $i$th-stage feature whose scale is $1/2^{i+1}$ of the input image, \ie $H_i=H/2^{i+1}$ and $W_i=W/2^{i+1}$. We denote $\{X_i\}_{i=1}^4$ as the mapped features by lateral connections where $X_i \in \mathbb{R}^{C \times H_i \times W_i}$ and $C$ is empirically set to $256$. $\{F_i\}_{i=1}^4$ represent the progressively aggregated features from top down, where $F_4=\mathcal{G}(X_4)$ and $\forall i=1, 2, 3,  F_i=\mathcal{F}_i(F_{i+1}, X_i)$. $\mathcal{G}$ is defined as identity mapping in this study but can be any context module like ASPP~\cite{chen2018encoder}. $F_i$ corresponds to the CFT that is described in detail as follows.

\subsubsection{Category Feature Embedding}
Given the current aggregated feature $F_{i+1}$, we first obtain class masks $M_{i+1}$ and aligned features $\hat{F}_{i+1}$ after a normalization operation:
\begin{equation}
	\tilde{F}_{i+1}=Norm(F_{i+1}),
	\label{eq:1}
\end{equation}
\begin{equation}
	M_{i+1}=\Phi_{i+1}^m(\tilde{F}_{i+1}),~\hat{F}_{i+1}=\Phi_{i+1}^f(\tilde{F}_{i+1}),
	\label{eq:2}
\end{equation}
where $\Phi_{i+1}^m$ and $\Phi_{i+1}^f$ are two linear transformations implemented by $1 \times 1$ convolutional layers without activation. $M_{i+1} \in \mathbb{R}^{L \times H_{i+1} \times W_{i+1}}$ and $L$ is the number of semantic categories. $M_{i+1}$ is then reshaped to $\mathbb{R}^{L \times N_{i+1}}$ where $N_{i+1}=H_{i+1} \times W_{i+1}$. For each mask $m_{i+1, l} \in \mathbb{R}^{N_{i+1}}$ of category $l \in [1, L]$ in $M_{i+1}$, we apply \emph{softmax} (denoted by $\sigma$) to obtain attention weights, following Eq.~\ref{eq:3}.
\begin{equation}
	\hat{M}_{i+1}=\sigma(M_{i+1})=
	\left[
	\begin{matrix}
		\sigma(m_{i+1, 1}) \\
		\cdots\\
		\sigma(m_{i+1, L})
	\end{matrix}
	\right].
	\label{eq:3}
\end{equation}
$\hat{F}_{i+1}$ is reshaped to $\mathbb{R}^{C \times N_{i+1}}$, then we calculate the unified feature embedding for each semantic category at the $i$th stage by Eq.~\ref{eq:4}.
\begin{equation}
	\mathcal{J}_i=\hat{M}_{i+1} (\hat{F}_{i+1})^T, \mathcal{J}_i \in \mathbb{R}^{L \times C}.
	\label{eq:4}
\end{equation}
In this way, for each category $l$, we define its unified feature embedding $\mathcal{J}_{i, l} \in \mathbb{R}^C$ at the current aggregation stage as the weighted average of all pixels belonging to that category. Therefore, a precise mask prediction matters as it serves as a coarse indicator for the pixel's expected semantic category.

\subsubsection{Category Feature Transformation}
Once we have the learned unified category feature embeddings $\mathcal{J}_i$ that retain rich semantics at the $i$th aggregation stage, we explore the capacity of multi-head attention to adaptively transfer them to each feature dot in $X_i$ that directly comes from the backbone network (see Fig.~\ref{fig:2} and Fig.~\ref{fig:3}).

First, we briefly review the multi-head attention mechanism. It uses three basic elements of $query$, $key$ and $value$ as the inputs that are denoted as $Q$, $K$ and $V$, respectively.
\begin{equation}
	\mathit{MHA}(Q, K, V)=\left[\sigma\left(\frac{Q_h{K_h}^T}{\sqrt{d_h}}\right)V_h\right]_{h=1}^\mathcal{H}W^O,
	\label{eq:5}
\end{equation}
where $Q=\left[Q_h\right]_{h=1}^\mathcal{H}$, $K=\left[K_h\right]_{h=1}^\mathcal{H}$ and $V=\left[V_h\right]_{h=1}^\mathcal{H}$. $[\cdot]$ represents concatenation operation. $\mathcal{H}$ means the number of heads and $d_h$ the scaling factor. $W^O$ is a learnable matrix.

In the proposed category feature transformation, we compute $\mathit{MHA}(Q_i, K_i, V_i)$ to interact with the $i$th-stage feature $X_i$, where $Q_i$, $K_i$ and $V_i$ are calculated as follows:
\begin{equation}
	Q_i=Norm(X_i)W_i^Q,~K_i=\mathcal{J}_iW_i^K,~V_i=\mathcal{J}_iW_i^V.
	\label{eq:6}
\end{equation}
$W_i^Q$, $W_i^K$ and $W_i^V$ are all learnable parameters. Finally, we obtain the aggregated feature $F_i$ as follows:
\begin{equation}
	\breve{F}_i=\mathit{MHA}(Q_i, K_i, V_i)+X_i,
	\label{eq:7}
\end{equation}
\begin{equation}
	F_i=\mathit{FFN}(Norm(\breve{F}_i))+\breve{F}_i,
	\label{eq:8}
\end{equation}
where FFN denotes a feed-forward network. As we observe no apparent accuracy perturbation in our experiments when introducing the relative positional encoding~\cite{bello2019attention} or sinusoidal embedding~\cite{parmar2018image}, we follow Xie~\etal~\cite{xie2021segformer} to implement FFN with depth-wise separable convolutions~\cite{howard2017mobilenets}, resulting in a positional encoding-free architecture.

\subsection{Loss Function}
The overall loss function is defined as Eq.~\ref{eq:9}, where $\mathcal{L}_\mathit{ce}$ means cross-entropy (CE) loss for final pixel classification. We aslo use a mask loss that is a linear combination of focal loss~\cite{lin2017focal} and dice loss~\cite{milletari2016v} to supervise the learning of category masks, as in~\cite{zhang2022segvit}. Masks are summed orderly for computing the loss. $\lambda_d$ and $\lambda_f$ are empirically set to 2 and 5.
\begin{equation}
	\mathcal{L}=\mathcal{L}_\mathit{ce}+\underbrace{\lambda_d\mathcal{L}_\mathit{dice}+\lambda_f\mathcal{L}_\mathit{focal}}_{\mathcal{L}_\mathit{mask}}.
	\label{eq:9}
\end{equation}

\section{Experiments}
\label{sec:4}
\subsection{Experimental Settings}
\subsubsection{Datasets and Metrics}
\textbf{ADE20K}~\cite{zhou2017scene} contains about $20k$ images for training and $2k$ images for validation with $150$ semantic categories. \textbf{Cityscapes}~\cite{cordts2016cityscapes} is a complex street scene parsing dataset with $19$ semantic categories, containing $5k$ finely labeled images and $20k$ coarsely labeled images. In this study, we only use finely labeled data, including $2,975/500/1,525$ images for training/validation/testing. \textbf{Pascal-Context}~\cite{mottaghi2014role} is a dataset with $4,996$ images for training and $5,104$ images for validation. It has $60$ labeled semantic categories, one of which is \emph{background}. 

Following standard convention, we adopt the mean intersection over union (mIoU) to evaluate the segmentation accuracy. The number of model parameters (\#Params) and float-point operations (FLOPs) estimated by \emph{fvcore}\footnote{\url{https://github.com/facebookresearch/fvcore}} are used to assess the model complexity.

\subsubsection{Implementation Details}
The backbone network is pre-trained on ImageNet~\cite{russakovsky2015imagenet} and the newly added modules are randomly initialized. During training, we employ the poly learning rate scheduler, \ie $lr=baselr \times \left(1-iter/total\_iter\right)^{power}$ where $power=1.0$. We employ AdamW~\cite{loshchilov2017decoupled} as the optimizer and train the model for $80k$ iterations on Pascal-Context and $160k$ iterations on the others. The batch size is set to $8$ for Cityscapes and $16$ for the others. For data augmentation, we follow the general pipelines, including random scale between $0.5 \times \sim 2.0 \times$, random cropping, random horizontal flip with a probability of $0.5$ and photometric distortion. The crop size is set to $512^2$ or $640^2$ for ADE20K, $480^2$ for Pascal-Context and $1024^2$ for Cityscapes. A sliding window of $1024^2$ is adopted for inference on Cityscapes.


\subsection{Ablation Study}
We conduct ablation experiments using ADE20K dataset, expecting to provide more insights into the proposed CFT. The backbone network is ResNet-50~\cite{he2016deep} and MiT-B2~\cite{xie2021segformer}. The crop size is $512^2$, and all ablation results are based on single-scale inputs.

\begin{table}[t]
    \centering
    \resizebox{.98\linewidth}{!}{
    \begin{tabular}{lcc|l}
		\toprule
		\textbf{Method} & \textbf{\#Params} & \textbf{FLOPs} & \textbf{mIoU(\%)} \\
		\midrule\midrule
		FCN~\cite{long2015fully} & 28.9M & 24.3G & 33.15 \\
		SemFPN~\cite{kirillov2019panoptic} & 28.5M & 45.9G & 37.49 (+4.34) \\
		VT-FPN~\cite{wu2020visual} & 50.0M & 46.6G & 37.68 (+4.53) \\
		FaPN~\cite{huang2021fapn} & 31.4M & 59.3G & 39.57 (+6.42)  \\
		SegFormer (MLP)~\cite{xie2021segformer} & 24.8M & 29.9G & 32.37 (-0.78) \\
		AlignSeg~\cite{huang2022alignseg} & 34.7M & 85.8G & 39.96 (+6.81) \\
		\rowcolor{LightCyan} CFT (ours) & 29.5M & 81.0G & \textbf{43.00} (+9.85) \\
		\midrule
		FCN~\cite{long2015fully} & 26.0M & 21.0G & 44.17 \\
		SemFPN~\cite{kirillov2019panoptic} & 28.5M & 42.0G & 46.30 (+2.13) \\
		VT-FPN~\cite{wu2020visual} & 33.6M & 27.3G & 46.74 (+2.57) \\
		FaPN~\cite{huang2021fapn} & 30.1M & 55.4G & 47.31 (+3.14)  \\
		SegFormer (MLP)~\cite{xie2021segformer} & 24.8M & 26.0G & 46.53 (+2.36) \\
		AlignSeg~\cite{huang2022alignseg} & 30.7M & 81.0G & 47.33 (+3.16) \\
		\rowcolor{LightCyan} CFT (ours) & 29.5M & 77.1G & \textbf{48.80} (+4.63) \\
		\bottomrule
    \end{tabular}}
    \caption{Comparisons with other multi-stage feature aggregation methods, where FCN is employed for an impartial comparison. The upper part uses ResNet-50 as the backbone and the lower MiT-B2. FLOPs is estimated using an image size of $512^2$.}
    \label{table:1}
\end{table}

\begin{table}[t]
    \centering
    \resizebox{.95\columnwidth}{!}{
    \begin{tabular}{lcc|l}
        \toprule
        \textbf{Version} & \textbf{FFN} & \textbf{FLOPs} & \textbf{mIoU(\%)} \\
		\midrule\midrule
		SemFPN~\cite{kirillov2019panoptic} & - & 42.0G & 46.30 \\
        \midrule
		Naive & \checkmark & 112G & 47.17 (+0.87) \\
		AvgPool & \checkmark & 78.3G & 47.45 (+1.15) \\
		Unified category feature & \checkmark & 77.1G & \textbf{48.80} (+2.50) \\
		Unified category feature & $\times$ & 65.6G & 47.79 (+1.49) \\
		\bottomrule
    \end{tabular}}
    \caption{Ablation results for key components. The backbone network is MiT-B2 and SemFPN is listed as the baseline.}
    \label{table:2}
\end{table}

\subsubsection{Multi-stage Feature Aggregation}
We verify the effectiveness of the proposed CFT framework for multi-stage feature aggregation, which is the fundamental insight of this study. We report comparisons with both CNN and Transformer backbones in Tab.~\ref{table:1}. FCN~\cite{long2015fully} only leverages the final stage feature, and others fully utilize multi-stage features. SemFPN~\cite{kirillov2019panoptic} adopts the naive FPN structure~\cite{lin2017feature} for semantic segmentation, suggesting the significance of multi-stage feature aggregation. VT-FPN~\cite{wu2020visual}, FaPN~\cite{huang2021fapn} and AlignSeg~\cite{huang2022alignseg} are enhanced FPN variants that further improve segmentation accuracy. We also notice that some methods fail to generalize to various backbones, \eg SegFormer~\cite{xie2021segformer}. Our proposed CFT, in contrast, obtains superior performance with both CNN and vision transformer backbones while preventing an increase in model complexity, \eg lower than the most recent AlignSeg~\cite{huang2022alignseg} in both model parameters and FLOPs.

\begin{figure}[t]
    \centering
    \includegraphics[width=0.96\columnwidth]{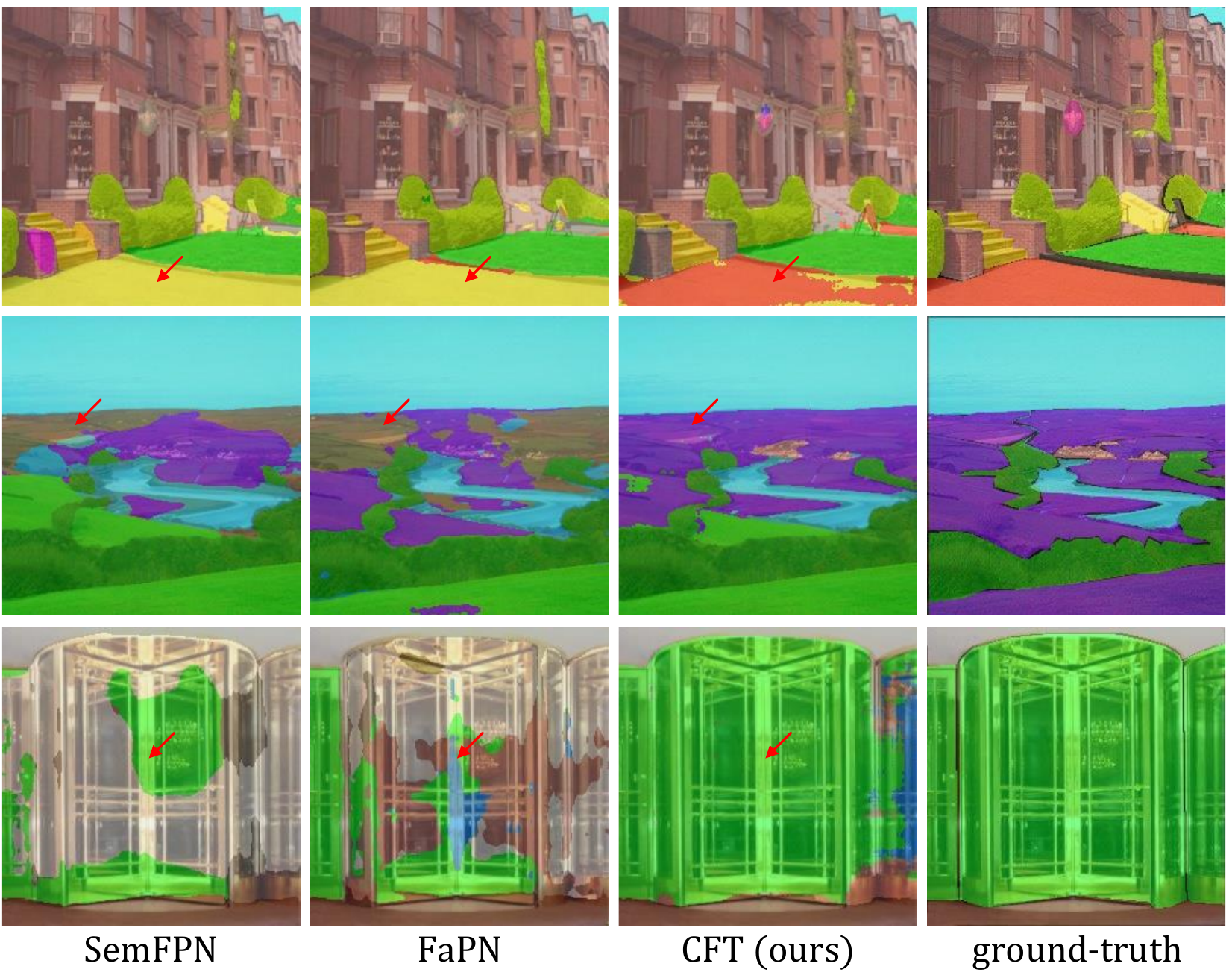}
    \caption{Visual comparisons of different top-down feature aggregation methods on ADE20K validation set. Red arrows indicate significant contrasts. The backbone is MiT-B2.}
    \label{fig:4}
\end{figure}

We further plot the distribution of per-category accuracy gain after aggregating multi-stage features in Fig.~\ref{fig:1}, from which we can observe that CFT effectively alleviates performance drop in minority categories and obtains a more consistent accuracy improvement in all categories, suggesting a more robust multi-stage feature aggregation paradigm. Some visual comparisons regarding the top-down feature aggregation regime are juxtaposed in Fig.~\ref{fig:4}.

\subsubsection{Verification of the Key Insight}
Previous methods~\cite{huang2021fapn,huang2022alignseg} show that a context misalignment issue exists in feature aggregation with the conventional bilinear upsampling and point-wise summation. The direct intuition of this work is to employ the multi-head attention mechanism for adjacent-stage feature aggregation because of the powerful context modeling capacity and the potential of \emph{implicit feature reconstruction}. To this end, a simple implementation uses the high-level feature $F_{i+1}$ as key/value and the corresponding low-level feature $X_i$ as query. As shown in Tab.~\ref{table:2}, this naive version achieves significant performance gain ($0.87\%$ mIoU$\uparrow$) over the SemFPN~\cite{kirillov2019panoptic} baseline, suggesting the effectiveness. However, the computation cost increases (FLOPs $42.0$G $\rightarrow 112$G), which is unendurable when scaling to large models.

To reduce computation, we follow an efficient design to downsample key/value by average pooling\footnote{We experiment with bilinear downsampling, average pooling and stride convolution and find that the average pooling can work well in our experiments.}, which gathers features in regular grids (\ie pooling windows) and surprisingly results in an improved segmentation accuracy of $47.45\%$ mIoU, showing that computing the similarity score between any pairwise pixel feature dots is unnecessary in the top down aggregation process. We further propose to learn a unified feature embedding for each semantic category to ensure consistent semantics, which boosts the performance by a large margin to $48.80\%$ mIoU. FFN also plays a significant role in adjacent-stage feature fusion as omitting it leads to a notable accuracy drop (mIoU $48.80\% \rightarrow 47.79\%$). Overall, these results evidence the capacity of multi-head attention and unified category embeddings for aggregating multi-stage features without an explicit upsampling operation.

\subsubsection{Verification of the Mask Loss}
We observed that appropriate mask supervision is vital for category-embedding-based methods, \eg OCRNet~\cite{yuan2020object} and CondNet~\cite{yu2021condnet}. We conduct verification from this aspect. Tab.~\ref{table:3} gives the results, where we stick to the same dataset and training settings as the original paper for a fair comparison. The results suggest that appropriate mask supervision varies among methods, and the valuable accuracy improvement of CFT does not come from a unique mask loss but from the novel architecture itself. 

\begin{figure}[t]
    \centering
    \includegraphics[width=0.98\columnwidth]{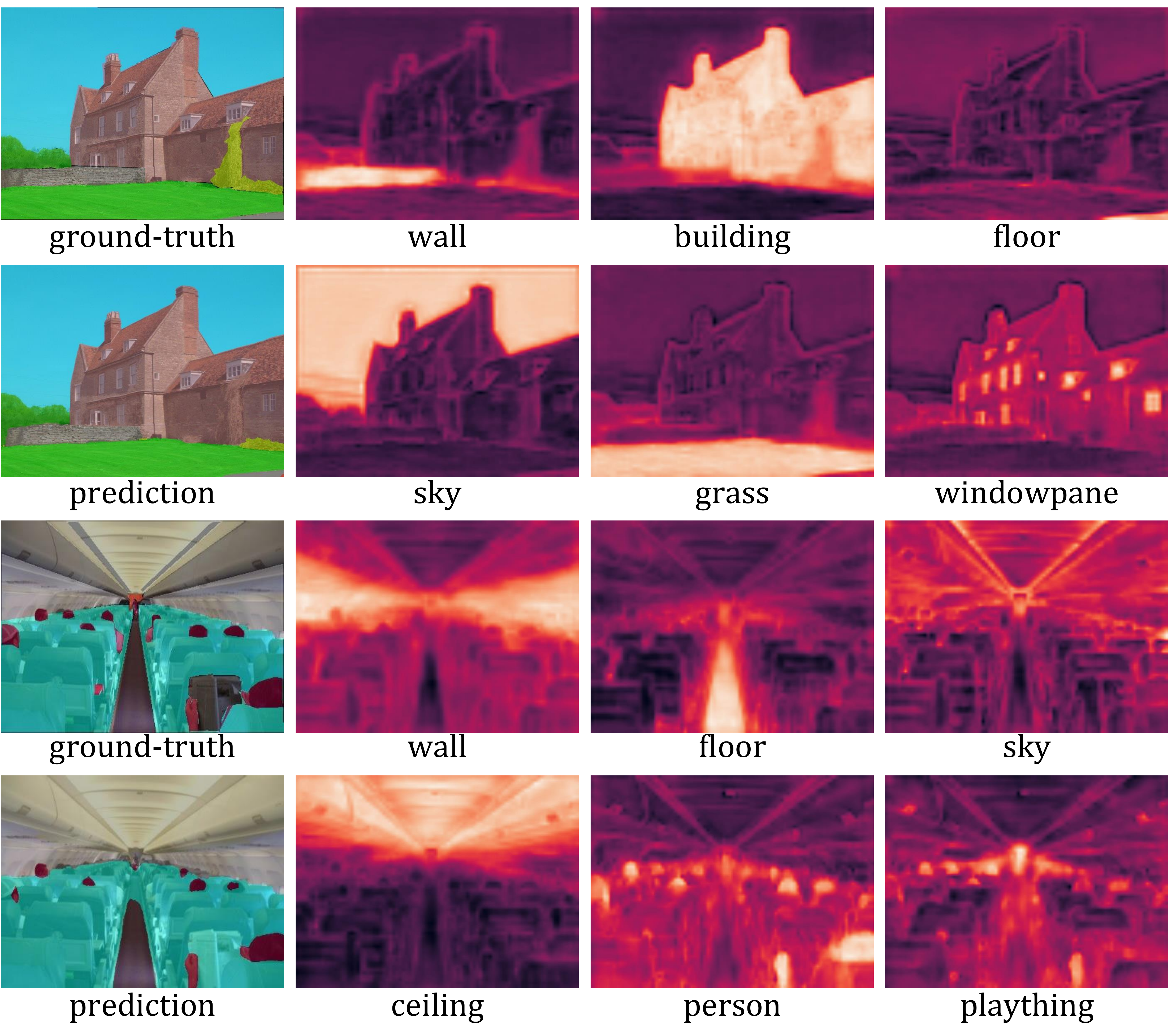}
    \caption{Visualization of class masks. The last column lists masks regarding irrelevant categories that do not appear in the image.}
    \label{fig:5}
\end{figure}

We visualize examples of learned class masks in Fig.~\ref{fig:5}, where we list masks regarding both relevant and irrelevant semantic categories. As we can see, with the help of an appropriate mask loss, CFT learns precise masks that identify each semantic region. Since relevant semantics are well addressed while irrelevant semantics are effectively suppressed by these masks, accurate category feature embeddings can be obtained, contributing to higher segmentation accuracy. Meanwhile, an intriguing phenomenon suggests that meaningful patterns naturally emerge even without valid annotations, \eg the \emph{windowpane} at the end of the second row.

\begin{table}[t]
    \centering
    \small
    \resizebox{.99\columnwidth}{!}{
    \begin{tabular}{l|c|c}
        \toprule
        \textbf{Method} & \textbf{Dataset} & \textbf{mIoU(\%)} \\
         \midrule\midrule
         OCRNet~\cite{yuan2020object} & Cityscapes & 77.31 \\
         \quad + cross-entropy loss $\clubsuit$ & Cityscapes & \underline{79.58} \\
         \quad + our mask loss $\mathcal{L}_\mathit{mask}$ & Cityscapes & \textbf{79.61} \\
         \midrule
         CondNet~\cite{yu2021condnet} & ADE20K & 41.75 \\
         \quad + dice loss $\clubsuit$ & ADE20K & \textbf{42.37} \\
         \quad + cross-entropy loss & ADE20K & 40.43 \\
         \quad + our mask loss $\mathcal{L}_\mathit{mask}$ & ADE20K & 40.90 \\
         \midrule
         CFT (ours) & ADE20K & 46.96 \\
         \quad + cross-entropy loss & ADE20K & 46.91 \\
         \quad + our mask loss $\mathcal{L}_\mathit{mask}$ & ADE20K & \textbf{48.80} \\
         \bottomrule
    \end{tabular}}
    \caption{Ablation for different mask loss. $\clubsuit$ indicates the setting of the original paper.}
    \label{table:3}
\end{table}

\begin{figure*}[t]
    \centering
    \includegraphics[width=0.92\textwidth]{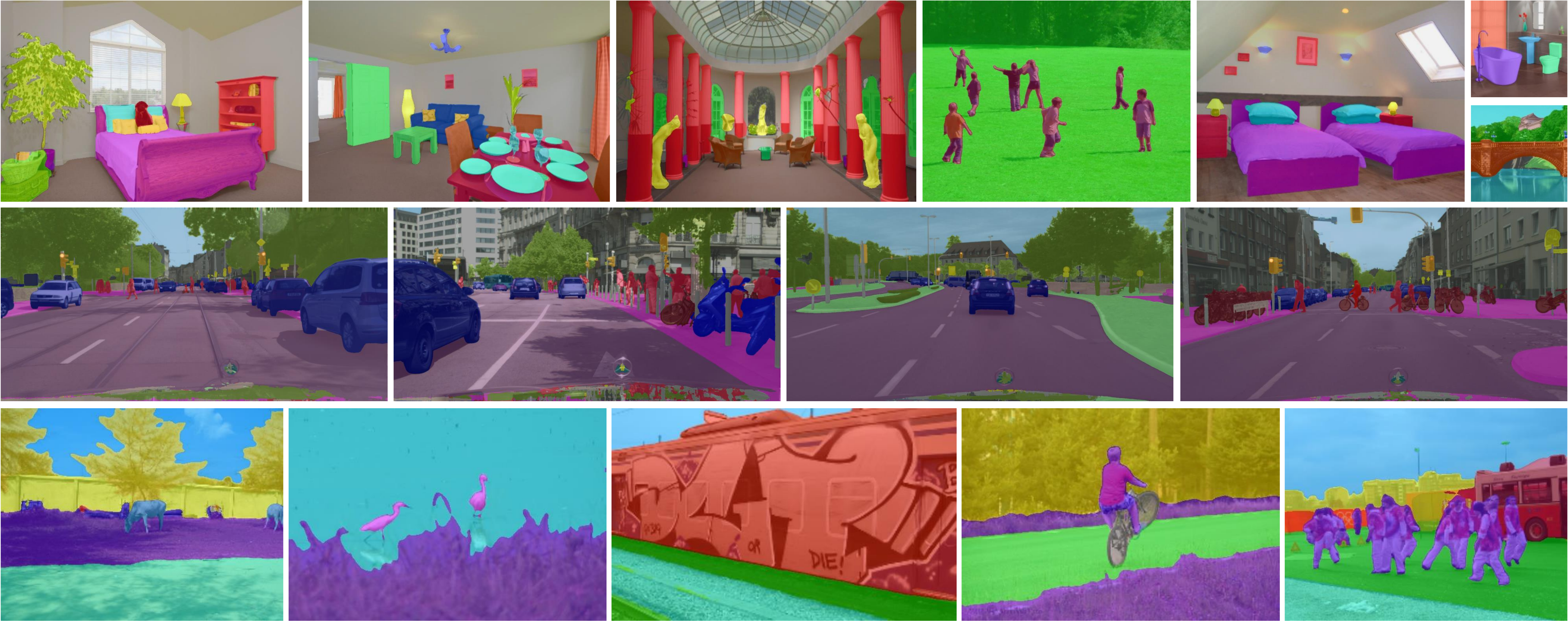}
    \caption{Visualized predictions on ADE20K, Cityscapes and Pascal-Context validation set. The backbone is SwinT-Large.}
    \label{fig:6}
\end{figure*}

\subsection{Results on ADE20K}
We report numerical comparisons with various backbones in Tab~\ref{table:4}. In the CNN backbone camp, CFT obtains superior segmentation accuracy with \emph{minimal} model parameters and computational costs based on the common ResNet-101~\cite{he2016deep}. As for vision transformer backbones, CFT gains a significant accuracy improvement ($1.2\%$ mIoU$\uparrow$) over SegFormer~\cite{xie2021segformer} with the same MiT-B5 backbone. When working with the more powerful SwinT~\cite{liu2021swin}, CFT yields $55.1\%$ mIoU with \emph{only} $200$M parameters and $417$G FLOPs, outperforming all other counterparts. For instance, CFT obtains superior and comparable performance to the most recent ViT-Adapter~\cite{chen2023vision} and SegViT~\cite{zhang2022segvit} with much fewer model parameters and FLOPs. Furthermore, we show that employing the proposed CFT as the pixel decoder in the Mask2Former framework~\cite{cheng2022masked} boosts the performance from $57.3\%$ to $58.5\%$ mIoU, which validates its effectiveness as an off-the-shelf plug-in component. CFT has shown robustness and scalability to various backbone networks. Visualized predictions are shown in Fig.~\ref{fig:6}.

\begin{table}[t]
    \begin{center}
    \resizebox{1.\columnwidth}{!}{
    \begin{tabular}{lccc|r}
		\toprule
		\textbf{Method} & \textbf{Backbone} & \textbf{\#Params} & \textbf{FLOPs} & \textbf{mIoU(\%)} \\
		\midrule\midrule
		FCN~\cite{long2015fully} & ResNet-101 & 66.2M & 276G & 39.9/41.4 \\
		PSPNet~\cite{zhao2017pyramid} & ResNet-101 & 65.7M & 257G & 44.4/45.4 \\
		EncNet~\cite{zhang2018context} & ResNet-101 & 52.7M & 219G & 42.6/44.0 \\
		Non-local~\cite{wang2018non} & ResNet-101 & 66.7M & 287G & 44.6/45.8 \\
		CCNet~\cite{huang2019ccnet} & ResNet-101 & 66.5M & 279G & 43.7/45.0 \\
		ANN~\cite{zhu2019asymmetric} & ResNet-101 & 62.9M & 264G & -/45.2 \\
		OCRNet~\cite{yuan2020object} & ResNet-101 & 55.5M & 231G & -/45.3 \\
		AlignSeg~\cite{huang2022alignseg} & ResNet-101 & 53.7M & 105G & -/46.0 \\
        \rowcolor{LightCyan} CFT (ours) & ResNet-101 & 48.5M & 100G & 45.0/\textbf{46.8} \\
        \midrule
        SegFormer~\cite{xie2021segformer} & MiT-B5 & 82.0M & 126G & 51.0/51.8 \\
        NRD~\cite{zhang2021dynamic} & MiT-B5 & 99.4M & 74.7G & 51.2/51.9 \\
        \rowcolor{LightCyan} CFT (ours) & MiT-B5 & 86.7M & 206G & 52.2/\textbf{53.0} \\
		\midrule
		SETR-PUP~\cite{zheng2021rethinking} & ViT-Large & 308M & 711G & 48.2/50.0 \\
		Segmenter~\cite{strudel2021segmenter} & ViT-Large & 333M & 672G & 51.7/53.6 \\
		StructToken-CSE~\cite{lin2023structtoken} & ViT-Large$\dagger$ & 350M & \textgreater700G & 52.8/54.2 \\
        ViT-Adapter~\cite{chen2023vision} & ViT-Large$\dagger$ & 364M & - & 53.4/54.4 \\
        SegViT~\cite{zhang2022segvit} & ViT-Large$\dagger$ & 328M & 638G & -/\textbf{55.2} \\
        KNet~\cite{zhang2021k} & SwinT-Large$\dagger$ & 245M & 659G & -/54.3 \\	
		SenFormer~\cite{bousselham2022efficient} & SwinT-Large$\dagger$ & 233M & 546G & 53.1/54.2 \\
		\rowcolor{LightCyan} CFT (ours) & SwinT-Large$\dagger$ & 200M & 417G & 53.5/55.1 \\
        \midrule
		Mask2Former~\cite{cheng2022masked} & SwinT-Large$\dagger$ & 215M & 403G & 56.1/57.3 \\
        Mask2Former + FaPN~\cite{huang2021fapn} & SwinT-Large$\dagger$ & 217M & - & 56.4/57.7 \\
		\rowcolor{LightCyan} Mask2Former + CFT (ours) & SwinT-Large$\dagger$ & 216M & 438G & 57.4/\textbf{58.5} \\
        \bottomrule
    \end{tabular}}
    \end{center}
    \caption{Results on ADE20K validation set with single- and multi-scale inputs given by $\cdot/\cdot$. FLOPs is estimated using an image size of $512^2$ for CNN backbones and $640^2$ for Transformer ones. Backbones pre-trained with ImageNet-22K are marked with $\dagger$.}
    \label{table:4}
\end{table}

\subsection{Results on Cityscapes}
Tab.~\ref{table:5} shows the results on Cityscapes dataset, including comparisons on both validation and test subsets. We use only the \emph{train-fine} images to train the model. The proposed CFT obtains the optimal $84.4\%$ and $82.9\%$ mIoU on two subsets, respectively. CFT is surprisingly superior to Mask2Former~\cite{cheng2022masked} without the complicated mask decoding process. We also present comparisons with the most recent kMaX-DeepLab~\cite{yu2022k}, where CFT obtains comparable accuracy. We note that kMaX-DeepLab is an improved design based on Mask2Former, which already incorporates an enhanced FPN variant inside. And its adopted backbone ConvNeXt~\cite{liu2022convnet} is believed to be more potent as it obtains higher accuracy on ImageNet classification. When evaluating on the test set, using extra \emph{val-fine} images for training further increases the accuracy to $83.3\%$ mIoU. Note that we use no training skills like online hard examples mining~\cite{wu2016high} or auxiliary losses~\cite{zhao2017pyramid} on multiple scales. 

\begin{table}[t]
    \begin{center}
    \resizebox{1.\columnwidth}{!}{
    \begin{tabular}{lcc|c}
        \toprule
		\textbf{Method} & \textbf{Backbone} & \textbf{Subset} & \textbf{mIoU(\%)} \\
		\midrule\midrule
		PSPNet~\cite{zhao2017pyramid} & ResNet-101 & val & 78.5 \\
		CCNet~\cite{huang2019ccnet} & ResNet-101 & val & 81.3 \\
		CANet~\cite{tang2022compensating} & ResNet-101 & val & 81.9 \\
        OCRNet~\cite{yuan2020object} & ResNet-101 & val & 81.8 \\
		AlignSeg~\cite{huang2022alignseg} & ResNet-101 & val & 82.4 \\
		SegFormer~\cite{xie2021segformer} & MiT-B5 & val & 84.0 \\
		SETR-PUP~\cite{zheng2021rethinking} & ViT-Large & val & 82.2 \\
		Segmenter~\cite{strudel2021segmenter} & ViT-Large & val & 81.3 \\
		StructToken-PWE~\cite{lin2023structtoken} & ViT-Large & val & 82.1 \\
		Mask2Former~\cite{cheng2022masked} & SwinT-Large$\dagger$ & val & 84.3 \\
		\rowcolor{LightCyan} CFT (ours) & SwinT-Large$\dagger$ & val & \textbf{84.4} \\
        \midrule
    $\clubsuit$kMaX-DeepLab~\cite{yu2022k} & ConvNeXt-L$\dagger$ & val & 83.5 \\
        \rowcolor{LightCyan} $\clubsuit$CFT (ours) & SwinT-Large$\dagger$ & val & 83.5 \\
		\midrule
        $\ddagger$ACFNet & ResNet-101 & test & 81.8 \\
		CCNet~\cite{huang2019ccnet} & ResNet-101 & test & 81.9 \\
		AlignSeg~\cite{huang2022alignseg} & ResNet-101 & test & 81.5 \\
		SETR-PUP~\cite{zheng2021rethinking} & ViT-Large & test & 81.0 \\
		SegFormer~\cite{xie2021segformer} & MiT-B5 & test & 82.2 \\
		\rowcolor{LightCyan} CFT (ours) & SwinT-Large$\dagger$ & test & \textbf{82.9} \\
		\rowcolor{LightCyan} $\ddagger$CFT (ours) & SwinT-Large$\dagger$ & test & \textbf{83.3} \\
		\bottomrule
    \end{tabular}}
    \end{center}
    \caption{Results on Cityscapes dataset with multi-scale inference, where $\clubsuit$ indicates single-scale results. Backbones pre-trained with ImageNet-22K are marked with $\dagger$. $\ddagger$ means training with \emph{trainval-fine} images.}
    \label{table:5}
\end{table}

\subsection{Results on Pascal-Context}
Tab.~\ref{table:6} shows the results on Pascal-Context dataset. For a clear comparison, we follow SegViT~\cite{zhang2022segvit} to evaluate our method and report scores regarding both $59$ classes (excluding \emph{background}) and $60$ classes (including \emph{background}). The proposed CFT reaches $65.6\%$ and $59.5\%$ mIoU respectively that outperforms all other methods. Notably, CFT is superior to the previous optimal SegViT~\cite{zhang2022segvit} on this benchmark. 

\begin{table}[t]
    \begin{center}
    \resizebox{.99\columnwidth}{!}{
    \begin{tabular}{lc|cc}
    \toprule
    \textbf{Method} & \textbf{Backbone} & \textbf{mIoU$_{59}$(\%)} & \textbf{mIoU$_{60}$(\%)} \\
        \midrule\midrule
		PSPNet~\cite{zhao2017pyramid} & ResNet-101 & 47.8 & - \\
		CPNet~\cite{yu2020context} & ResNet-101 & 53.9 & - \\
        OCRNet~\cite{yuan2020object} & ResNet-101 & 54.8 & - \\
		NRD~\cite{zhang2021dynamic} & ResNet-101 & 54.1 & 49.0 \\
        CondNet~\cite{yu2021condnet} & ResNet-101 & 56.0 & - \\
		CAA~\cite{huang2022channelized} & ResNet-101 & 55.0 & - \\
		SETR-MLA~\cite{zheng2021rethinking} & ViT-Large & - & 55.8 \\
		Segmenter~\cite{strudel2021segmenter} & ViT-Large & - & 59.0 \\
        SegViT~\cite{zhang2022segvit} & ViT-Large$\dagger$ & 65.3 & 59.3 \\
		CAR~\cite{huang2022car} & SwinT-Large$\dagger$ & 59.0 & - \\
		SenFormer~\cite{bousselham2022efficient} & SwinT-Large$\dagger$ & 64.5 & - \\
		\rowcolor{LightCyan} CFT (ours) & SwinT-Large$\dagger$ & \textbf{65.6} & \textbf{59.5} \\
		\bottomrule
    \end{tabular}}
    \end{center}
    \caption{Results on Pascal-Context validation set with multi-scale inference. Backbones pre-trained with ImageNet-22K are marked with $\dagger$.}
    \label{table:6}
\end{table}

\section{Conclusion}
This study explores the capacity of the prevalent multi-head attention mechanism for multi-stage feature aggregation in semantic image segmentation. We show its effectiveness in dynamically fusing two adjacent-stage features. We introduce a novel module with explicit learning via a mask loss to learn unified feature embeddings for each semantic category, ensuring consistent semantics by following natural object structure in the real world and reducing computational complexity. Extensive experimental studies suggest that the proposed CFT framework provides a robust top-down feature aggregation solution for semantic segmentation. We also show that CFT can work as an off-the-shelf component for existing methods with feature pyramids. In the future, we expect to exploit its ability as a potent FPN variant to address other dense prediction tasks and open vocabulary segmentation by using non-parametric clustering to obtain category embeddings.

\bibliography{egbib_aaai}

\begin{thebibliography}{55}
\providecommand{\natexlab}[1]{#1}

\bibitem[{Ba, Kiros, and Hinton(2016)}]{ba2016layer}
Ba, J.~L.; Kiros, J.~R.; and Hinton, G.~E. 2016.
\newblock Layer normalization.
\newblock \emph{arXiv preprint arXiv:1607.06450}.

\bibitem[{Bello et~al.(2019)Bello, Zoph, Vaswani, Shlens, and
  Le}]{bello2019attention}
Bello, I.; Zoph, B.; Vaswani, A.; Shlens, J.; and Le, Q.~V. 2019.
\newblock Attention augmented convolutional networks.
\newblock In \emph{Proceedings of the IEEE/CVF international conference on
  computer vision}, 3286--3295.

\bibitem[{Bousselham et~al.(2022)Bousselham, Thibault, Pagano, Machireddy,
  Gray, Chang, and Song}]{bousselham2022efficient}
Bousselham, W.; Thibault, G.; Pagano, L.; Machireddy, A.; Gray, J.; Chang,
  Y.~H.; and Song, X. 2022.
\newblock Efficient Self-Ensemble for Semantic Segmentation.
\newblock \emph{The British Machine Vision Conference}.

\bibitem[{Chen et~al.(2017)Chen, Papandreou, Schroff, and
  Adam}]{chen2017rethinking}
Chen, L.-C.; Papandreou, G.; Schroff, F.; and Adam, H. 2017.
\newblock Rethinking atrous convolution for semantic image segmentation.
\newblock \emph{arXiv preprint arXiv:1706.05587}.

\bibitem[{Chen et~al.(2018)Chen, Zhu, Papandreou, Schroff, and
  Adam}]{chen2018encoder}
Chen, L.-C.; Zhu, Y.; Papandreou, G.; Schroff, F.; and Adam, H. 2018.
\newblock Encoder-decoder with atrous separable convolution for semantic image
  segmentation.
\newblock In \emph{Proceedings of the European conference on computer vision
  (ECCV)}, 801--818.

\bibitem[{Chen et~al.(2023)Chen, Duan, Wang, He, Lu, Dai, and
  Qiao}]{chen2023vision}
Chen, Z.; Duan, Y.; Wang, W.; He, J.; Lu, T.; Dai, J.; and Qiao, Y. 2023.
\newblock Vision Transformer Adapter for Dense Predictions.
\newblock \emph{International Conference on Learning Representations}.

\bibitem[{Cheng et~al.(2022)Cheng, Misra, Schwing, Kirillov, and
  Girdhar}]{cheng2022masked}
Cheng, B.; Misra, I.; Schwing, A.~G.; Kirillov, A.; and Girdhar, R. 2022.
\newblock Masked-attention mask transformer for universal image segmentation.
\newblock In \emph{Proceedings of the IEEE/CVF conference on computer vision
  and pattern recognition}, 1290--1299.

\bibitem[{Cheng, Schwing, and Kirillov(2021)}]{cheng2021per}
Cheng, B.; Schwing, A.; and Kirillov, A. 2021.
\newblock Per-pixel classification is not all you need for semantic
  segmentation.
\newblock \emph{Advances in Neural Information Processing Systems}, 34:
  17864--17875.

\bibitem[{Cordts et~al.(2016)Cordts, Omran, Ramos, Rehfeld, Enzweiler,
  Benenson, Franke, Roth, and Schiele}]{cordts2016cityscapes}
Cordts, M.; Omran, M.; Ramos, S.; Rehfeld, T.; Enzweiler, M.; Benenson, R.;
  Franke, U.; Roth, S.; and Schiele, B. 2016.
\newblock The cityscapes dataset for semantic urban scene understanding.
\newblock In \emph{Proceedings of the IEEE conference on computer vision and
  pattern recognition}, 3213--3223.

\bibitem[{Dosovitskiy et~al.(2020)Dosovitskiy, Beyer, Kolesnikov, Weissenborn,
  Zhai, Unterthiner, Dehghani, Minderer, Heigold, Gelly
  et~al.}]{dosovitskiy2020image}
Dosovitskiy, A.; Beyer, L.; Kolesnikov, A.; Weissenborn, D.; Zhai, X.;
  Unterthiner, T.; Dehghani, M.; Minderer, M.; Heigold, G.; Gelly, S.; et~al.
  2020.
\newblock An Image is Worth 16x16 Words: Transformers for Image Recognition at
  Scale.
\newblock In \emph{International Conference on Learning Representations}.

\bibitem[{He et~al.(2016)He, Zhang, Ren, and Sun}]{he2016deep}
He, K.; Zhang, X.; Ren, S.; and Sun, J. 2016.
\newblock Deep residual learning for image recognition.
\newblock In \emph{Proceedings of the IEEE conference on computer vision and
  pattern recognition}, 770--778.

\bibitem[{Howard et~al.(2017)Howard, Zhu, Chen, Kalenichenko, Wang, Weyand,
  Andreetto, and Adam}]{howard2017mobilenets}
Howard, A.~G.; Zhu, M.; Chen, B.; Kalenichenko, D.; Wang, W.; Weyand, T.;
  Andreetto, M.; and Adam, H. 2017.
\newblock Mobilenets: Efficient convolutional neural networks for mobile vision
  applications.
\newblock \emph{arXiv preprint arXiv:1704.04861}.

\bibitem[{Huang et~al.(2021)Huang, Lu, Cheng, and He}]{huang2021fapn}
Huang, S.; Lu, Z.; Cheng, R.; and He, C. 2021.
\newblock FaPN: Feature-aligned pyramid network for dense image prediction.
\newblock In \emph{Proceedings of the IEEE/CVF International Conference on
  Computer Vision}, 864--873.

\bibitem[{Huang et~al.(2022{\natexlab{a}})Huang, Kang, Chen, Zhe, Jia, Bao, and
  He}]{huang2022car}
Huang, Y.; Kang, D.; Chen, L.; Zhe, X.; Jia, W.; Bao, L.; and He, X.
  2022{\natexlab{a}}.
\newblock Car: Class-aware regularizations for semantic segmentation.
\newblock In \emph{European Conference on Computer Vision}, 518--534. Springer.

\bibitem[{Huang et~al.(2022{\natexlab{b}})Huang, Kang, Jia, Liu, and
  He}]{huang2022channelized}
Huang, Y.; Kang, D.; Jia, W.; Liu, L.; and He, X. 2022{\natexlab{b}}.
\newblock Channelized Axial Attention--Considering Channel Relation within
  Spatial Attention for Semantic Segmentation.
\newblock In \emph{Proceedings of the AAAI Conference on Artificial
  Intelligence}, volume~36, 1016--1025.

\bibitem[{Huang et~al.(2019)Huang, Wang, Huang, Huang, Wei, and
  Liu}]{huang2019ccnet}
Huang, Z.; Wang, X.; Huang, L.; Huang, C.; Wei, Y.; and Liu, W. 2019.
\newblock Ccnet: Criss-cross attention for semantic segmentation.
\newblock In \emph{Proceedings of the IEEE/CVF international conference on
  computer vision}, 603--612.

\bibitem[{Huang et~al.(2022{\natexlab{c}})Huang, Wei, Wang, Liu, Huang, and
  Shi}]{huang2022alignseg}
Huang, Z.; Wei, Y.; Wang, X.; Liu, W.; Huang, T.~S.; and Shi, H.
  2022{\natexlab{c}}.
\newblock AlignSeg: Feature-Aligned Segmentation Networks.
\newblock \emph{IEEE Transactions on Pattern Analysis and Machine
  Intelligence}, 44(1): 550--557.

\bibitem[{Ioffe and Szegedy(2015)}]{ioffe2015batch}
Ioffe, S.; and Szegedy, C. 2015.
\newblock Batch normalization: Accelerating deep network training by reducing
  internal covariate shift.
\newblock In \emph{International conference on machine learning}, 448--456.
  pmlr.

\bibitem[{Kirillov et~al.(2019)Kirillov, Girshick, He, and
  Doll{\'a}r}]{kirillov2019panoptic}
Kirillov, A.; Girshick, R.; He, K.; and Doll{\'a}r, P. 2019.
\newblock Panoptic feature pyramid networks.
\newblock In \emph{Proceedings of the IEEE/CVF conference on computer vision
  and pattern recognition}, 6399--6408.

\bibitem[{Li et~al.(2022)Li, Mao, Girshick, and He}]{li2022exploring}
Li, Y.; Mao, H.; Girshick, R.; and He, K. 2022.
\newblock Exploring plain vision transformer backbones for object detection.
\newblock In \emph{European Conference on Computer Vision}, 280--296. Springer.

\bibitem[{Lin et~al.(2023)Lin, Liang, Wu, He, Chen, and
  Tian}]{lin2023structtoken}
Lin, F.; Liang, Z.; Wu, S.; He, J.; Chen, K.; and Tian, S. 2023.
\newblock Structtoken: Rethinking semantic segmentation with structural prior.
\newblock \emph{IEEE Transactions on Circuits and Systems for Video
  Technology}.

\bibitem[{Lin et~al.(2017{\natexlab{a}})Lin, Doll{\'a}r, Girshick, He,
  Hariharan, and Belongie}]{lin2017feature}
Lin, T.-Y.; Doll{\'a}r, P.; Girshick, R.; He, K.; Hariharan, B.; and Belongie,
  S. 2017{\natexlab{a}}.
\newblock Feature pyramid networks for object detection.
\newblock In \emph{Proceedings of the IEEE conference on computer vision and
  pattern recognition}, 2117--2125.

\bibitem[{Lin et~al.(2017{\natexlab{b}})Lin, Goyal, Girshick, He, and
  Doll{\'a}r}]{lin2017focal}
Lin, T.-Y.; Goyal, P.; Girshick, R.; He, K.; and Doll{\'a}r, P.
  2017{\natexlab{b}}.
\newblock Focal loss for dense object detection.
\newblock In \emph{Proceedings of the IEEE international conference on computer
  vision}, 2980--2988.

\bibitem[{Liu et~al.(2021)Liu, Lin, Cao, Hu, Wei, Zhang, Lin, and
  Guo}]{liu2021swin}
Liu, Z.; Lin, Y.; Cao, Y.; Hu, H.; Wei, Y.; Zhang, Z.; Lin, S.; and Guo, B.
  2021.
\newblock Swin transformer: Hierarchical vision transformer using shifted
  windows.
\newblock In \emph{Proceedings of the IEEE/CVF international conference on
  computer vision}, 10012--10022.

\bibitem[{Liu et~al.(2022)Liu, Mao, Wu, Feichtenhofer, Darrell, and
  Xie}]{liu2022convnet}
Liu, Z.; Mao, H.; Wu, C.-Y.; Feichtenhofer, C.; Darrell, T.; and Xie, S. 2022.
\newblock A convnet for the 2020s.
\newblock In \emph{Proceedings of the IEEE/CVF conference on computer vision
  and pattern recognition}, 11976--11986.

\bibitem[{Long, Shelhamer, and Darrell(2015)}]{long2015fully}
Long, J.; Shelhamer, E.; and Darrell, T. 2015.
\newblock Fully convolutional networks for semantic segmentation.
\newblock In \emph{Proceedings of the IEEE conference on computer vision and
  pattern recognition}, 3431--3440.

\bibitem[{Loshchilov and Hutter(2017)}]{loshchilov2017decoupled}
Loshchilov, I.; and Hutter, F. 2017.
\newblock Decoupled weight decay regularization.
\newblock \emph{arXiv preprint arXiv:1711.05101}.

\bibitem[{Milletari, Navab, and Ahmadi(2016)}]{milletari2016v}
Milletari, F.; Navab, N.; and Ahmadi, S.-A. 2016.
\newblock V-net: Fully convolutional neural networks for volumetric medical
  image segmentation.
\newblock In \emph{2016 fourth international conference on 3D vision (3DV)},
  565--571. Ieee.

\bibitem[{Mottaghi et~al.(2014)Mottaghi, Chen, Liu, Cho, Lee, Fidler, Urtasun,
  and Yuille}]{mottaghi2014role}
Mottaghi, R.; Chen, X.; Liu, X.; Cho, N.-G.; Lee, S.-W.; Fidler, S.; Urtasun,
  R.; and Yuille, A. 2014.
\newblock The role of context for object detection and semantic segmentation in
  the wild.
\newblock In \emph{Proceedings of the IEEE conference on computer vision and
  pattern recognition}, 891--898.

\bibitem[{Pang et~al.(2019)Pang, Li, Shen, and Shao}]{pang2019towards}
Pang, Y.; Li, Y.; Shen, J.; and Shao, L. 2019.
\newblock Towards bridging semantic gap to improve semantic segmentation.
\newblock In \emph{Proceedings of the IEEE/CVF International Conference on
  Computer Vision}, 4230--4239.

\bibitem[{Parmar et~al.(2018)Parmar, Vaswani, Uszkoreit, Kaiser, Shazeer, Ku,
  and Tran}]{parmar2018image}
Parmar, N.; Vaswani, A.; Uszkoreit, J.; Kaiser, L.; Shazeer, N.; Ku, A.; and
  Tran, D. 2018.
\newblock Image transformer.
\newblock In \emph{International conference on machine learning}, 4055--4064.
  PMLR.

\bibitem[{Russakovsky et~al.(2015)Russakovsky, Deng, Su, Krause, Satheesh, Ma,
  Huang, Karpathy, Khosla, Bernstein et~al.}]{russakovsky2015imagenet}
Russakovsky, O.; Deng, J.; Su, H.; Krause, J.; Satheesh, S.; Ma, S.; Huang, Z.;
  Karpathy, A.; Khosla, A.; Bernstein, M.; et~al. 2015.
\newblock Imagenet large scale visual recognition challenge.
\newblock \emph{International journal of computer vision}, 115: 211--252.

\bibitem[{Strudel et~al.(2021)Strudel, Garcia, Laptev, and
  Schmid}]{strudel2021segmenter}
Strudel, R.; Garcia, R.; Laptev, I.; and Schmid, C. 2021.
\newblock Segmenter: Transformer for semantic segmentation.
\newblock In \emph{Proceedings of the IEEE/CVF international conference on
  computer vision}, 7262--7272.

\bibitem[{Tang et~al.(2022)Tang, Liu, Zhang, Jiang, Zhang, Zhu, and
  Tang}]{tang2022compensating}
Tang, Q.; Liu, F.; Zhang, T.; Jiang, J.; Zhang, Y.; Zhu, B.; and Tang, X. 2022.
\newblock Compensating for local ambiguity with encoder-decoder in urban scene
  segmentation.
\newblock \emph{IEEE Transactions on Intelligent Transportation Systems},
  23(10): 19224--19235.

\bibitem[{Tian et~al.(2019)Tian, He, Shen, and Yan}]{tian2019decoders}
Tian, Z.; He, T.; Shen, C.; and Yan, Y. 2019.
\newblock Decoders matter for semantic segmentation: Data-dependent decoding
  enables flexible feature aggregation.
\newblock In \emph{Proceedings of the IEEE/CVF Conference on Computer Vision
  and Pattern Recognition}, 3126--3135.

\bibitem[{Vaswani et~al.(2017)Vaswani, Shazeer, Parmar, Uszkoreit, Jones,
  Gomez, Kaiser, and Polosukhin}]{vaswani2017attention}
Vaswani, A.; Shazeer, N.; Parmar, N.; Uszkoreit, J.; Jones, L.; Gomez, A.~N.;
  Kaiser, {\L}.; and Polosukhin, I. 2017.
\newblock Attention is all you need.
\newblock \emph{Advances in neural information processing systems}, 30.

\bibitem[{Wang et~al.(2018)Wang, Girshick, Gupta, and He}]{wang2018non}
Wang, X.; Girshick, R.; Gupta, A.; and He, K. 2018.
\newblock Non-local neural networks.
\newblock In \emph{Proceedings of the IEEE conference on computer vision and
  pattern recognition}, 7794--7803.

\bibitem[{Wu et~al.(2020)Wu, Xu, Dai, Wan, Zhang, Yan, Tomizuka, Gonzalez,
  Keutzer, and Vajda}]{wu2020visual}
Wu, B.; Xu, C.; Dai, X.; Wan, A.; Zhang, P.; Yan, Z.; Tomizuka, M.; Gonzalez,
  J.; Keutzer, K.; and Vajda, P. 2020.
\newblock Visual transformers: Token-based image representation and processing
  for computer vision.
\newblock \emph{arXiv preprint arXiv:2006.03677}.

\bibitem[{Wu, Shen, and Hengel(2016)}]{wu2016high}
Wu, Z.; Shen, C.; and Hengel, A. v.~d. 2016.
\newblock High-performance semantic segmentation using very deep fully
  convolutional networks.
\newblock \emph{arXiv preprint arXiv:1604.04339}.

\bibitem[{Xiao et~al.(2018)Xiao, Liu, Zhou, Jiang, and Sun}]{xiao2018unified}
Xiao, T.; Liu, Y.; Zhou, B.; Jiang, Y.; and Sun, J. 2018.
\newblock Unified perceptual parsing for scene understanding.
\newblock In \emph{Proceedings of the European conference on computer vision
  (ECCV)}, 418--434.

\bibitem[{Xie et~al.(2021)Xie, Wang, Yu, Anandkumar, Alvarez, and
  Luo}]{xie2021segformer}
Xie, E.; Wang, W.; Yu, Z.; Anandkumar, A.; Alvarez, J.~M.; and Luo, P. 2021.
\newblock SegFormer: Simple and efficient design for semantic segmentation with
  transformers.
\newblock \emph{Advances in Neural Information Processing Systems}, 34:
  12077--12090.

\bibitem[{Yu et~al.(2021)Yu, Shao, Gao, and Sang}]{yu2021condnet}
Yu, C.; Shao, Y.; Gao, C.; and Sang, N. 2021.
\newblock CondNet: Conditional classifier for scene segmentation.
\newblock \emph{IEEE Signal Processing Letters}, 28: 758--762.

\bibitem[{Yu et~al.(2020)Yu, Wang, Gao, Yu, Shen, and Sang}]{yu2020context}
Yu, C.; Wang, J.; Gao, C.; Yu, G.; Shen, C.; and Sang, N. 2020.
\newblock Context prior for scene segmentation.
\newblock In \emph{Proceedings of the IEEE/CVF conference on computer vision
  and pattern recognition}, 12416--12425.

\bibitem[{Yu et~al.(2022)Yu, Wang, Qiao, Collins, Zhu, Adam, Yuille, and
  Chen}]{yu2022k}
Yu, Q.; Wang, H.; Qiao, S.; Collins, M.; Zhu, Y.; Adam, H.; Yuille, A.; and
  Chen, L.-C. 2022.
\newblock k-means Mask Transformer.
\newblock In \emph{European Conference on Computer Vision}, 288--307. Springer.

\bibitem[{Yuan, Chen, and Wang(2020)}]{yuan2020object}
Yuan, Y.; Chen, X.; and Wang, J. 2020.
\newblock Object-contextual representations for semantic segmentation.
\newblock In \emph{Computer Vision--ECCV 2020: 16th European Conference,
  Glasgow, UK, August 23--28, 2020, Proceedings, Part VI 16}, 173--190.
  Springer.

\bibitem[{Zhang et~al.(2021{\natexlab{a}})Zhang, Tian, Shen
  et~al.}]{zhang2021dynamic}
Zhang, B.; Tian, Z.; Shen, C.; et~al. 2021{\natexlab{a}}.
\newblock Dynamic neural representational decoders for high-resolution semantic
  segmentation.
\newblock \emph{Advances in Neural Information Processing Systems}, 34:
  17388--17399.

\bibitem[{Zhang et~al.(2022)Zhang, Tian, Tang, Chu, Wei, Shen
  et~al.}]{zhang2022segvit}
Zhang, B.; Tian, Z.; Tang, Q.; Chu, X.; Wei, X.; Shen, C.; et~al. 2022.
\newblock Segvit: Semantic segmentation with plain vision transformers.
\newblock \emph{Advances in Neural Information Processing Systems}, 35:
  4971--4982.

\bibitem[{Zhang et~al.(2020)Zhang, Zhang, Tang, Wang, Hua, and
  Sun}]{zhang2020feature}
Zhang, D.; Zhang, H.; Tang, J.; Wang, M.; Hua, X.; and Sun, Q. 2020.
\newblock Feature pyramid transformer.
\newblock In \emph{Computer Vision--ECCV 2020: 16th European Conference,
  Glasgow, UK, August 23--28, 2020, Proceedings, Part XXVIII 16}, 323--339.
  Springer.

\bibitem[{Zhang et~al.(2019)Zhang, Chen, Li, Hong, Liu, Ma, Han, and
  Ding}]{zhang2019acfnet}
Zhang, F.; Chen, Y.; Li, Z.; Hong, Z.; Liu, J.; Ma, F.; Han, J.; and Ding, E.
  2019.
\newblock Acfnet: Attentional class feature network for semantic segmentation.
\newblock In \emph{Proceedings of the IEEE/CVF International Conference on
  Computer Vision}, 6798--6807.

\bibitem[{Zhang et~al.(2018)Zhang, Dana, Shi, Zhang, Wang, Tyagi, and
  Agrawal}]{zhang2018context}
Zhang, H.; Dana, K.; Shi, J.; Zhang, Z.; Wang, X.; Tyagi, A.; and Agrawal, A.
  2018.
\newblock Context encoding for semantic segmentation.
\newblock In \emph{Proceedings of the IEEE conference on Computer Vision and
  Pattern Recognition}, 7151--7160.

\bibitem[{Zhang et~al.(2021{\natexlab{b}})Zhang, Pang, Chen, and
  Loy}]{zhang2021k}
Zhang, W.; Pang, J.; Chen, K.; and Loy, C.~C. 2021{\natexlab{b}}.
\newblock K-net: Towards unified image segmentation.
\newblock \emph{Advances in Neural Information Processing Systems}, 34:
  10326--10338.

\bibitem[{Zhao et~al.(2017)Zhao, Shi, Qi, Wang, and Jia}]{zhao2017pyramid}
Zhao, H.; Shi, J.; Qi, X.; Wang, X.; and Jia, J. 2017.
\newblock Pyramid scene parsing network.
\newblock In \emph{Proceedings of the IEEE conference on computer vision and
  pattern recognition}, 2881--2890.

\bibitem[{Zheng et~al.(2021)Zheng, Lu, Zhao, Zhu, Luo, Wang, Fu, Feng, Xiang,
  Torr et~al.}]{zheng2021rethinking}
Zheng, S.; Lu, J.; Zhao, H.; Zhu, X.; Luo, Z.; Wang, Y.; Fu, Y.; Feng, J.;
  Xiang, T.; Torr, P.~H.; et~al. 2021.
\newblock Rethinking semantic segmentation from a sequence-to-sequence
  perspective with transformers.
\newblock In \emph{Proceedings of the IEEE/CVF conference on computer vision
  and pattern recognition}, 6881--6890.

\bibitem[{Zhou et~al.(2017)Zhou, Zhao, Puig, Fidler, Barriuso, and
  Torralba}]{zhou2017scene}
Zhou, B.; Zhao, H.; Puig, X.; Fidler, S.; Barriuso, A.; and Torralba, A. 2017.
\newblock Scene parsing through ade20k dataset.
\newblock In \emph{Proceedings of the IEEE conference on computer vision and
  pattern recognition}, 633--641.

\bibitem[{Zhu et~al.(2019)Zhu, Xu, Bai, Huang, and Bai}]{zhu2019asymmetric}
Zhu, Z.; Xu, M.; Bai, S.; Huang, T.; and Bai, X. 2019.
\newblock Asymmetric non-local neural networks for semantic segmentation.
\newblock In \emph{Proceedings of the IEEE/CVF international conference on
  computer vision}, 593--602.

\end{thebibliography}

\clearpage
\section{Supplementary}
\subsection{Experiments}
\subsubsection{Training Settings}
In CFT, we set both the number of attention heads $\mathcal{H}$ and the feature expansion ratio in FFN as $4$. For the \emph{AvgPool} version in Tab.~\ref{table:2}, we utilize the adaptive pooling layer to downsample key/value to the same size as $X_4^B$. We use layer normalization~\cite{ba2016layer} for all CFT blocks since we observe in multiple trials that it yields more stable performance even though batch normalization~\cite{ioffe2015batch} achieves the highest peak accuracy.

The default hyper-parameter setting during training is shown in Tab.~\ref{table:7}. For other configuration we follow the standard setting in \emph{mmsegmentation}\footnote{\url{https://github.com/open-mmlab/mmsegmentation}} toolbox. We note that training with MiT~\cite{xie2021segformer} as the backbone network is unstable, which is rather sensitive to random seeds.

\begin{table}[h]
    \begin{center}
    \small
    \resizebox{.98\linewidth}{!}{
    \begin{tabular}{ccccc}
        \toprule
		\textbf{Dataset} & \textbf{Backbone} & \textbf{Crop} & $\boldsymbol{\mathit{baselr}}$ & $\boldsymbol{wd}$ \\
		\midrule\midrule
		ADE20K & ResNet-50/101 & $512^2$ & $6e-5$ & $1e-5$ \\
		ADE20K & MiT-B2 & $512^2$ & $6e-5$ & $1e-2$ \\
		ADE20K & MiT-B5 & $640^2$ & $6e-5$ & $4e-2$ \\
		ADE20K & SwinT-Large & $640^2$ & $6e-5$ & $3e-2$ \\
		\midrule
		Cityscapes & SwinT-Large & $1024^2$ & $4e-5$ & $3e-2$ \\
		\midrule
		Pascal-Context & SwinT-Large & $480^2$ & $2e-5$ & $1e-2$ \\
		\bottomrule
    \end{tabular}}
    \end{center}
    \caption{Detailed training hyper-parameters.}
    \label{table:7}
\end{table}

\subsubsection{Feature Aggregation Variants}
\begin{figure}[t]
    \centering
    \includegraphics[width=0.95\linewidth]{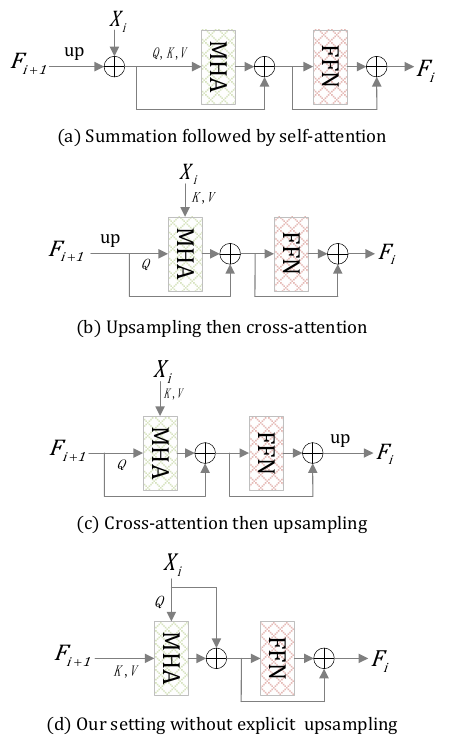}
    \caption{Variants of attention-based feature aggregation where $up$ means bilinear upsampling. We omit normalization, adaptive pooling and linear embedding layers for readability.}
    \label{fig:7}
\end{figure}

We study variants of attention-based feature aggregation, denoted by $@a$, $@b$ and $@c$, respectively. $\mathit{Variant}@a$ (see Fig.~\ref{fig:7}a) first upsamples $F_{i+1}$ to the same size as $X_i$. Self-attention is then applied on the summed-up feature map to capture global interactions, which enhances context representations for the next-layer aggregation. $\mathit{Variant}@b$ (see Fig.~\ref{fig:7}b) uses the upsampled $F_{i+1}$ as $Q$ and $X_i$ as $K/V$. In this setting, we can render higher-level semantics with visual attributes of lower-level pixels, according to~\cite{zhang2020feature}. $\mathit{Variant}@c$ (see Fig.~\ref{fig:7}c) differs from $\mathit{Variant}@b$ only in the position of the upsampling operator. Fig.~\ref{fig:7}d depicts the adopted setting in CFT, requiring no explicit upsampling operation. Note that we use adaptive pooling layers to downsample key/value in all variants to reduce memory footprint.

Results are reported in Tab.~\ref{table:8}, where SemFPN~\cite{kirillov2019panoptic} serves as the baseline. Both $\mathit{Variant}@a$ and our adopted setting achieve accuracy improvement over the baseline while $\mathit{Variant}@b$ and $@c$ show a notable performance degradation. We conjecture that noisy semantics reside in the bilinear upsampled $Q$ in $@b$ and $@c$ due to the constrained feature reconstruction capability, which has negative impacts on the attention-based context interaction and leads to inferior segmentation accuracy.

\begin{table}[t]
    \begin{center}	
    \small
    \resizebox{.99\linewidth}{!}{
    \begin{tabular}{lcc|l}
        \toprule
        \textbf{Variant} & \textbf{\#Params} & \textbf{FLOPs} & \textbf{mIoU(\%)} \\
        \midrule\midrule
		SemFPN~\cite{kirillov2019panoptic} & 28.46M & 42.01G & 46.30 \\
		$\mathit{Variant}@a$ & 29.27M & 80.49G & \underline{47.21} (+0.91) \\
		$\mathit{Variant}@b$ & 29.27M & 80.52G & 43.70 (-2.60) \\
		$\mathit{Variant}@c$ & 29.27M & 80.52G & 44.02 (-2.28) \\
		$\mathit{Variant}@d$ (ours) & 29.27M & 78.31G & \textbf{47.45} (+1.15) \\
    \bottomrule
    \end{tabular}}
    \end{center}
    \caption{Results of different attention-based feature aggregation variants on ADE20K validation set. The backbone netwotk is MiT-B2.}
    \label{table:8}
\end{table}

Learning category-specific feature embeddings that contain consistent semantics is another insight for a robust multi-stage feature aggregation paradigm. As these features serve as abstract context descriptors in the category granularity, learning them from high-level feature maps is preferable. This underpins the adopted setting (\ie Fig.~\ref{fig:7}d) apart from the implicit feature upsampling mode.

\subsubsection{Plain Backbone Network}
Recent works~\cite{li2022exploring} have demonstrated rich representation capacity of plain vision transformers. We thus explore the scalability of the proposed CFT on plain backbones, \eg the naive Vision Transformer (ViT)~\cite{dosovitskiy2020image}. Results compared with the widely adopted UPerNet~\cite{xiao2018unified} are shown in Tab.~\ref{table:9}. We select outputs of intermediate layers $\{3^{rd}, 6^{th}, 9^{th}\}$ along with that of the final layer to construct feature pyramids. We simply use bilinear interpolation followed by $3 \times 3$ convolution to rescale the features. Results suggest that CFT shows scalability to plain vision transformers.

\begin{table}[t]
    \begin{center}
    \small
    \resizebox{.95\linewidth}{!}{
    \begin{tabular}{lccc|c}
        \toprule
		\textbf{Method} & \textbf{Backbone} & \textbf{\#Params} & \textbf{FLOPs} & \textbf{mIoU(\%)} \\
		\midrule\midrule
		UPerNet & ViT-Base & 144M & 444G & 47.7/50.0 \\
		CFT (ours) & ViT-Base & 117M & 286G & \textbf{49.9}/\textbf{51.1} \\
		\bottomrule
    \end{tabular}}
    \end{center}
    \caption{Results on ADE20k dataset using plain backbone networks. $\cdot/\cdot$ indicates scores using single- and multi-scale inputs, respectively. The crop size is $512^2$.}
    \label{table:9}
\end{table}

\subsection{Visualization}
\subsubsection{Learned Masks}
As shown in Tab.~\ref{table:3}, incorporating an appropriate mask loss is critical to learn category-specific features by accurately detecting the corresponding irregular region, which we refer to as explicit learning. Implicit learning\footnote{Implicit learning here indicates that the model learns only with signals from the final predictions penalized by cross-entropy loss.} without mask loss leads to unsatisfactory performance for all category-attention-based methods. 

\begin{figure}[t]
    \centering
    \includegraphics[width=1.\linewidth]{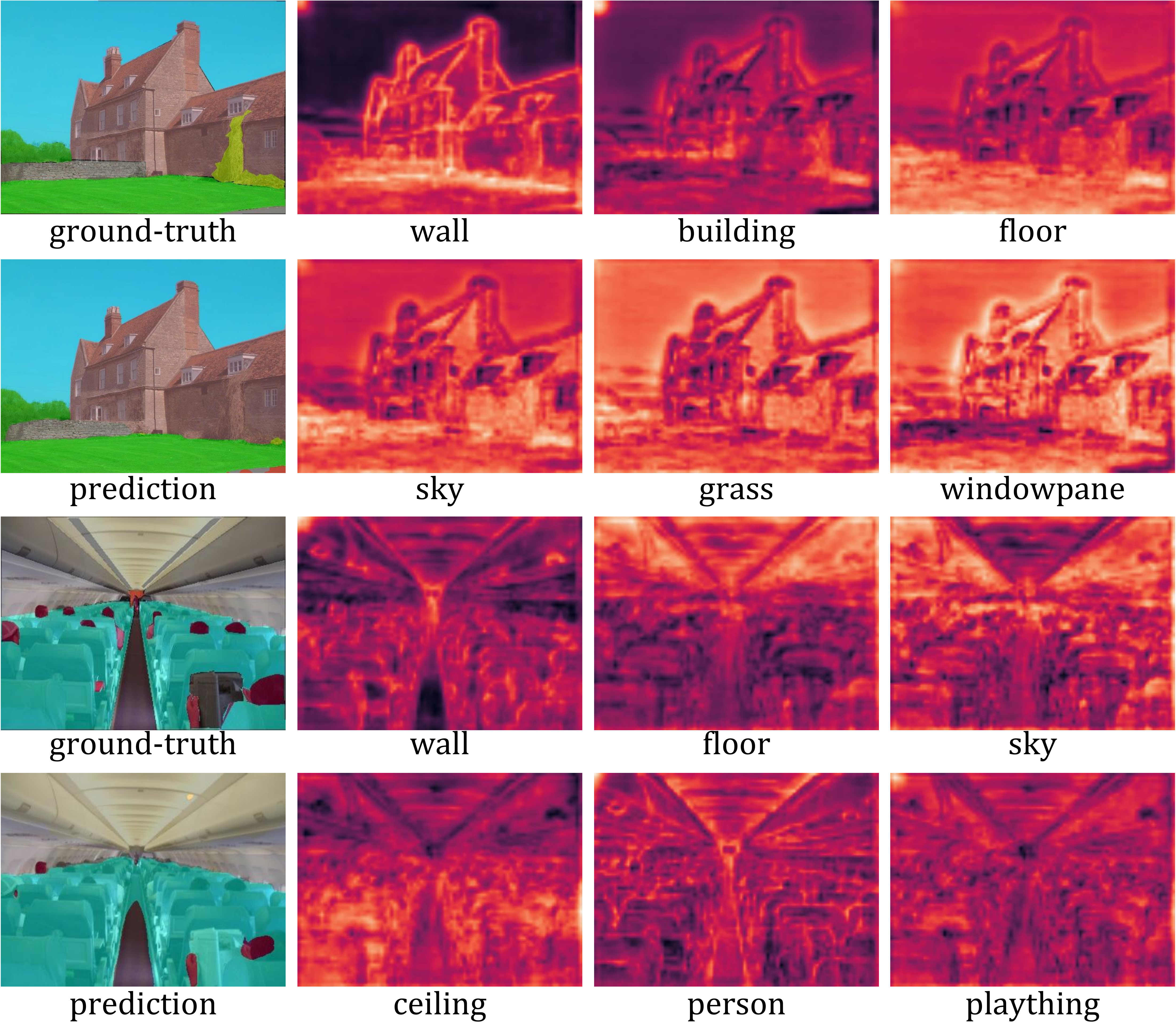}
    \caption{Visualization of class masks with implicit learning, \ie w/o the mask loss. The last column lists masks regarding irrelevant categories that do not appear in the image.}
    \label{fig:8}
\end{figure}

We present the learned masks with implicit learning in Fig.~\ref{fig:8}. No meaningful patterns show up, which stands in sharp contrast to Fig.~\ref{fig:5} with explicit learning. We note that although using an auxiliary loss has been widely accepted for quick convergence, it differs from our approach in some aspects. For example, previous works~\cite{zhao2017pyramid,zhang2019acfnet,zheng2021rethinking} use cross-entropy loss to penalize multiple-scale backbone features for prompting training, while CFT uses a mask loss to help learn a sharp localization mask in an explicit way.

\subsubsection{Visualized Predictions}
\begin{figure*}[t]
    \centering
    \includegraphics[width=0.97\linewidth]{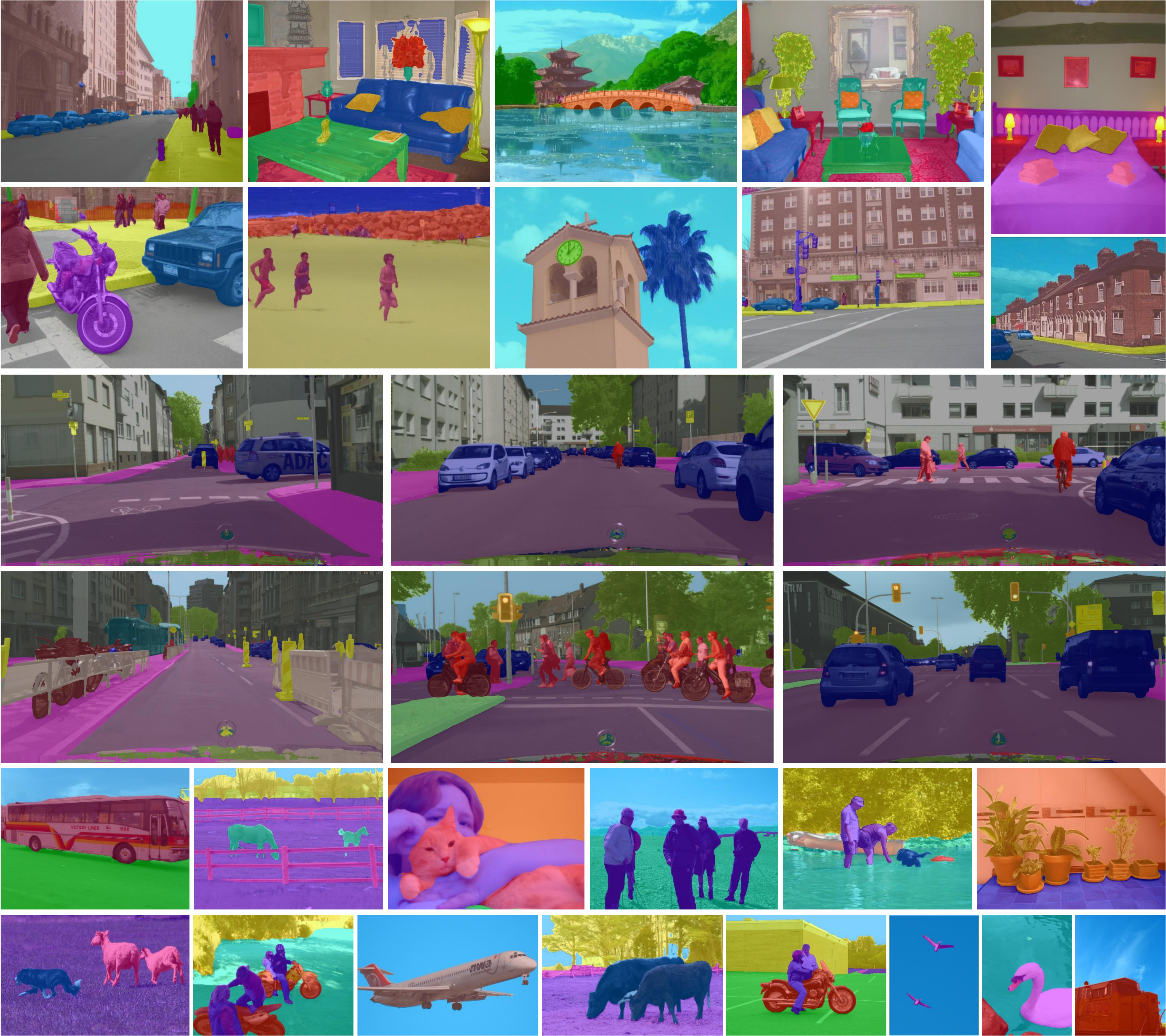}
    \caption{Competitive segmentation results on ADE20K, Cityscapes and Pascal-Context ($60$ categories) datasets. Best viewed in color.}
    \label{fig:9}
\end{figure*}

\begin{figure*}[t]
    \centering
    \includegraphics[width=0.88\linewidth]{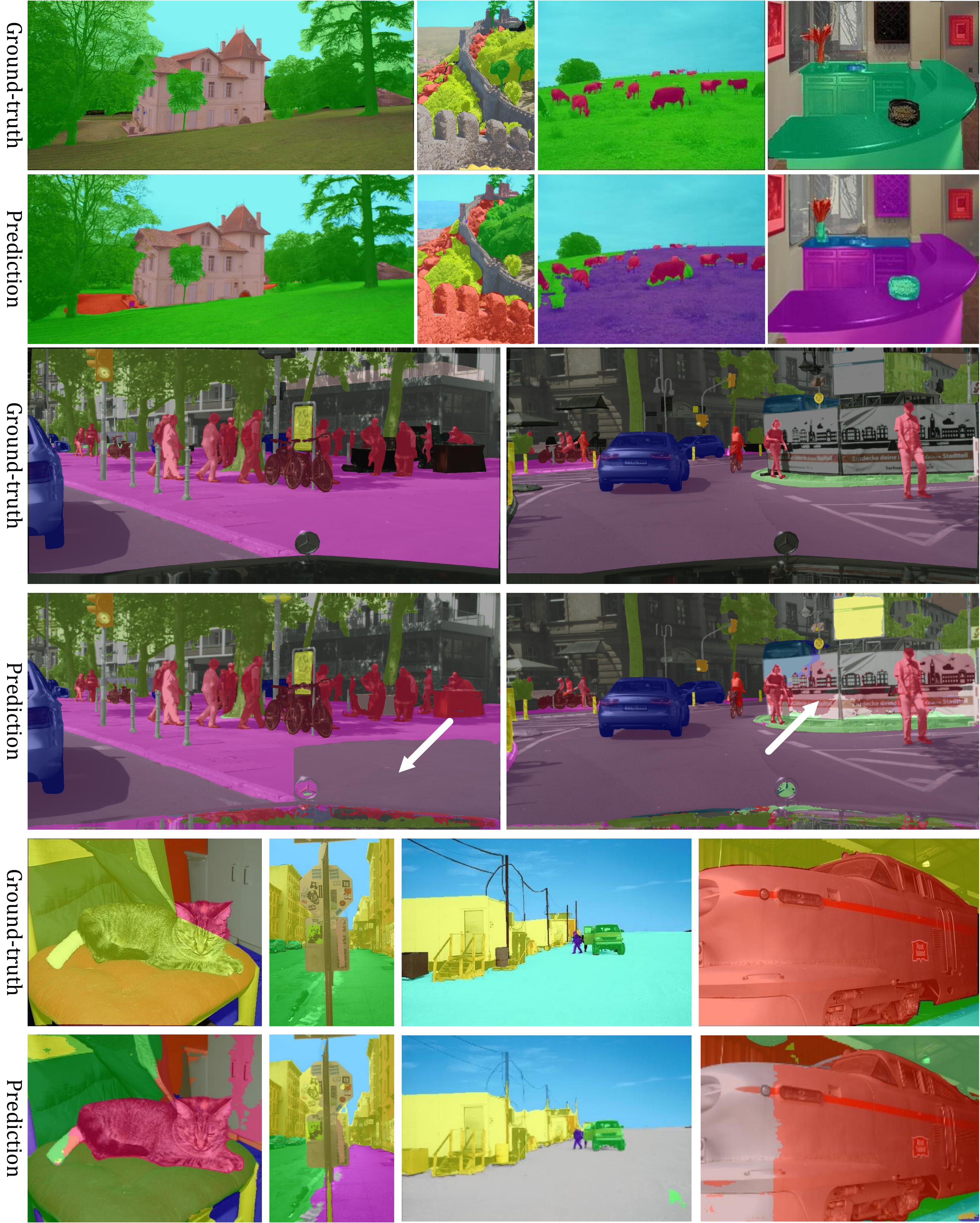}
    \caption{Failure cases on ADE20K, Cityscapes and Pascal-Context ($60$ categories) datasets. Best viewed in color.}
    \label{fig:10}
\end{figure*}

We present additional qualitative results including both competitive predictions and failure cases, as shown in Fig.~\ref{fig:9} and Fig.~\ref{fig:10}, respectively. We can observe that CFT produces sharp segmentation maps as it effectivelly utilizes multi-stage features (see Fig.~\ref{fig:9}). Meanwhile, different segments that have very similar appearance are confused (see Fig.~\ref{fig:10}).

\end{document}